# A Transformational Characterization of Markov Equivalence for Directed Acyclic Graphs with Latent Variables

**Jiji Zhang**
Department of Philosophy
Carnegie Mellon University
Pittsburgh, PA 15213
jiji@andrew.cmu.edu

**Peter Spirtes**
Department of Philosophy
Carnegie Mellon University
Institute for Human and Machine Cognition
University of West Florida
ps7z@andrew.cmu.edu

## Abstract

Different directed acyclic graphs (DAGs) may be Markov equivalent in the sense that they entail the same conditional independence relations among the observed variables. Chickering (1995) provided a transformational characterization of Markov equivalence for DAGs (with no latent variables), which is useful in deriving properties shared by Markov equivalent DAGs, and, with certain generalization, is needed to prove the asymptotic correctness of a search procedure over Markov equivalence classes, known as the GES algorithm.

For DAG models with latent variables, maximal ancestral graphs (MAGs) provide a neat representation that facilitates model search. However, no transformational characterization — analogous to Chickering's — of Markov equivalent MAGs is yet available. This paper establishes such a characterization for directed MAGs, which we expect will have similar uses as it does for DAGs.

## 1 INTRODUCTION

Markov equivalence between directed acyclic graphs (DAGs) has been characterized in several ways (e.g., Verma and Pearl 1990, Chickering 1995, Andersson et al. 1997). All of them have been found useful for various purposes. In particular, the transformational characterization provided by Chickering (1995) — that two Markov equivalent DAGs can be transformed to each other by a sequence of single edge reversals that preserve Markov equivalence — is useful in deriving properties shared by Markov equivalent DAGs. Moreover, when generalized, the transformational characterization implies the asymptotic correctness of the GES algorithm, an efficient search procedure over equivalence classes of DAGs (Meek 1996, Chickering 2002).

In many situations, however, we need also to consider DAGs with latent variables. Indeed there are cases where no DAGs can perfectly explain the observed conditional independence relations unless latent variables are introduced (see, e.g., Figure 2). But it is often undesirable to work with latent variable DAG models, especially with respect to model search. For example, given a set of observed variables, there are infinitely many latent variable DAG models to search over. Besides, to fit and score a DAG with latent variables is usually difficult due to statistical issues such as identifiability. Fortunately, such latent variable DAG models can be represented by ancestral graphical models (Richardson and Spirtes 2002), in that for any DAG with latent variables, there is a (maximal) ancestral graph that captures the exact observable conditional independence relations as well as some of the causal relations entailed by that DAG. Since ancestral graphs do not explicitly include latent variables, they are more amenable to search (Spirtes et al. 1997).

Markov equivalence for ancestral graphs has been characterized in ways analogous to the one given by Verma and Pearl (1990) for DAGs (Spirtes and Richardson 1996, Ali et al. 2004). However, no result is yet available that is analogous to Chickering's transformational characterization. In this paper we establish one for directed ancestral graphs. Specifically we show that two directed maximal ancestral graphs are Markov equivalent if and only if one can be transformed to the other by a sequence of single mark changes — adding or dropping an arrowhead — that preserve Markov equivalence. This characterization we expect will have similar uses as Chickering's does for DAGs. In particular, it is a step towards justifying the application of the GES algorithm to MAGs, and hence to latent variable DAG models.

The paper is organized as follows. The remainder of this section introduces the relevant definitions and no-tations. We then present the main result in section 2, drawing on some facts proved in Zhang and Spirtes (2005) and Ali et al. (2005). We conclude the paper in section 3 with a discussion of the potential application, limitation and generalization of our result.

## 1.1 DIRECTED ANCESTRAL GRAPHS

In full generality, an ancestral graph can contain three kinds of edges: directed edge ($\rightarrow$), bi-directed edge ($\leftrightarrow$) and undirected edge ($-$). In this paper, however, we will confine ourselves to directed ancestral graphs — which do not contain undirected edges — until section 3, where we explain why our result does not hold for general ancestral graphs. The class of directed ancestral graphs, due to its inclusion of bi-directed edges, is suitable for representing observed conditional independence structures in the presence of latent confounders (see Figure 2). Without undirected edges, however, ancestral graphs cannot represent the presence of latent selection variables.

By a **directed mixed graph** we denote an arbitrary graph that can have two kinds of edges: directed and bi-directed. The two ends of an edge we call **marks** or **orientations**. So the two marks of a bi-directed edge are both **arrowheads** ($>$), while a directed edge has one arrowhead and one **tail** ($-$) as its marks. Sometimes we say an edge is **into** (or **out of**) a vertex if the mark of the edge at the vertex is an arrowhead (or a tail). The meaning of the standard graph theoretical concepts, such as **parent/child**, **(directed) path**, **ancestor/descendant**, etc., remains the same in mixed graphs. Furthermore, if there is a bi-directed edge between two vertices $A$ and $B$ ($A \leftrightarrow B$), then $A$ is called a **spouse** of $B$ and $B$ a spouse of $A$.

**Definition 1 (ancestral).** *A directed mixed graph is **ancestral** if*

*(a1) there is no directed cycle; and*

*(a2) for any two vertices $A$ and $B$, if $A$ is a spouse of $B$ (i.e., $A \leftrightarrow B$), then $A$ is not an ancestor of $B$.*

Clearly DAGs are a special case of directed ancestral graphs (with no bi-directed edges). Condition (a1) is just the familiar one for DAGs. Condition (a2), together with (a1), defines a nice feature of arrowheads — that is, an arrowhead implies non-ancestorship. This motivates the term "ancestral" and induces a natural causal interpretation of ancestral graphs.

Mixed graphs encode conditional independence relations by essentially the same graphical criterion as the well-known *d-separation* for DAGs, except that in mixed graphs colliders can arise in more edge configurations than they do in DAGs. Given a path $u$ in a mixed graph, a non-endpoint vertex $V$ on $u$ is called a **collider** if the two edges incident to $V$ on $u$ are both into $V$, otherwise $V$ is called a **non-collider**.

**Definition 2 (m-separation).** *In a mixed graph, a path $u$ between vertices $A$ and $B$ is **active (m-connecting)** relative to a set of vertices $\mathbf{Z}$ ($A, B \notin \mathbf{Z}$) if*

  i. *every non-collider on $u$ is not a member of $\mathbf{Z}$;*

  ii. *every collider on $u$ is an ancestor of some member of $\mathbf{Z}$.*

*$A$ and $B$ are said to be **m-separated** by $\mathbf{Z}$ if there is no active path between $A$ and $B$ relative to $\mathbf{Z}$.*

The following property is true of DAGs: if two vertices are not adjacent, then there is a set of some other vertices that m-separates (d-separates) the two. This, however, is not true of directed ancestral graphs in general. For example, Figure 1(a) is an ancestral graph that fails this condition: $C$ and $D$ are not adjacent, but no subset of $\{A, B\}$ m-separates them.

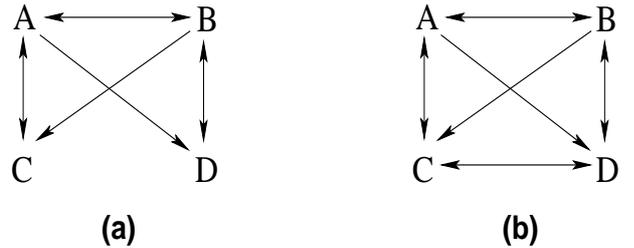

Figure 1: (a) an ancestral graph that is not maximal; (b) a maximal ancestral graph

This motivates the following definition:

**Definition 3 (maximality).** *A directed ancestral graph is said to be **maximal** if for any two non-adjacent vertices, there is a set of vertices that m-separates them.*

It is shown in Richardson and Spirtes (2002) that every non-maximal ancestral graph can be easily transformed to a *unique* supergraph that is ancestral and maximal by adding bi-directed edges. This justifies considering only those ancestral graphs that are maximal (MAGs). From now on, we focus on directed maximal ancestral graphs, which we will refer to as DMAGs. A notion closely related to maximality is that of inducing path:

**Definition 4 (inducing path).** *In an ancestral graph, a path $u$ between $A$ and $B$ is called an **inducing path** if every non-endpoint vertex on $u$ is a collider and is an ancestor of either $A$ or $B$.*

For example, in Figure 1(a), the path $\langle C, A, B, D \rangle$ is an inducing path between $C$ and $D$. Richardson and Spirtes (2002) proved that the presence of an inducing path is necessary and sufficient for two vertices not to be m-separated by any set. So, to show that a graph is maximal, it suffices to demonstrate that there is no inducing path between any two non-adjacent vertices in the graph.

Given any DAG with (or without) latent variables, the conditional independence relations as well as the causal relations among the observed variables can be represented by a DMAG that includes only the observed variables. The DMAG is constructed as follows: for every pair of observed variables, $O_i$ and $O_j$, put an edge between them if and only if they are not d-separated by any set of other observed variables in the given DAG, and mark an arrowhead at $O_i$ ($O_j$) on the edge if it is *not* an ancestor of $O_j$ ($O_i$) in the given DAG.

For example, Figure 2(a) is a DAG with latent variables $\{L1, L2, L3\}$. Figure 2(b) depicts the DMAG (G1) resulting from the above construction. The m-separation relations in G1 correspond exactly to the d-separation relations over $\{X1, X2, X3, X4, X5\}$ in Figure 2(a). By contrast, no DAG without extra latent variables has the exact same d-separation relations. Furthermore, the orientations in G1 accurately represent the ancestor relationships — which, upon natural interpretations, are causal relationships — among the observed variables in 2(a). (This, however, is not the case with G2.)

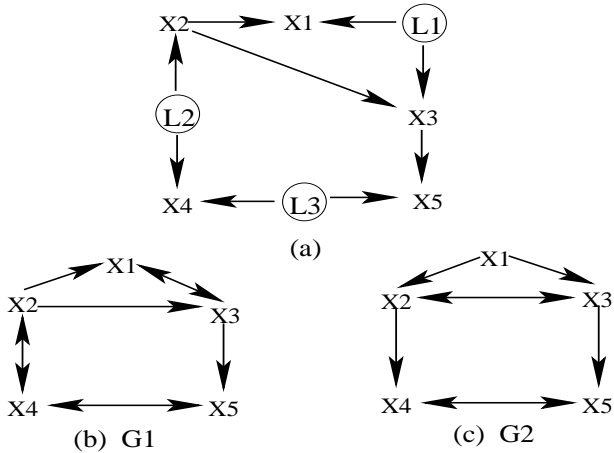

Figure 2: (a): A DAG with latent variables; (b): A DMAG that captures both the conditional independence and causal relations among the observed variables represented by (a); (c): A DMAG that entails the right conditional independence relations but not the right causal relations in (a).

## 1.2 MARKOV EQUIVALENCE

A DMAG represents the set of joint distributions that satisfy its *global Markov property*, i.e., the set of distributions of which the conditional independence relations entailed by m-separations in the DMAG hold. Hence, if two DMAGs share the same m-separation structures, then they represent the same set of distributions.

**Definition 5** (Markov equivalence). *Two DMAGs $\mathcal{G}_1, \mathcal{G}_2$ (with the same set of vertices) are **Markov equivalent** if for any three disjoint sets of vertices $\mathbf{X}, \mathbf{Y}, \mathbf{Z}$, $\mathbf{X}$ and $\mathbf{Y}$ are m-separated by $\mathbf{Z}$ in $\mathcal{G}_1$ if and only if $\mathbf{X}$ and $\mathbf{Y}$ are m-separated by $\mathbf{Z}$ in $\mathcal{G}_2$.*

Figure 2(c), for example, is a DMAG Markov equivalent to 2(b). It is well known that two DAGs are Markov equivalent if and only if they have the same adjacencies and the same unshielded colliders (Verma and Pearl 1990). (A triple $\langle A, B, C \rangle$ is said to be **unshielded** if $A, B$ are adjacent, $B, C$ are adjacent but $A, C$ are not adjacent.) The conditions are still necessary for Markov equivalence between DMAGs, but are not sufficient. For two DMAGs to be equivalent, some shielded colliders have to be present in both or neither of the graphs. The next definition is related to this.

**Definition 6** (discriminating path). *In a DMAG, a path between $X$ and $Y$, $u = \langle X, \cdots, W, V, Y \rangle$, is a **discriminating path** for $V$ if*

i. *$u$ includes at least three edges (i.e., at least four vertices as specified);*

ii. *$V$ is adjacent to an endpoint $Y$ on $u$; and*

iii. *$X$ is not adjacent to $Y$, and every vertex between $X$ and $V$ is a collider on $u$ and is a parent of $Y$.*

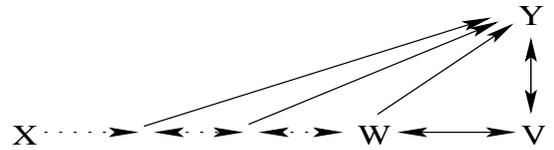

Figure 3: A discriminating path for $V$: the triple $\langle W, V, Y \rangle$ is "discriminated" to be a collider here.

Discriminating paths behave similarly to unshielded triples in that if $u = \langle X, \cdots, W, V, Y \rangle$ is discriminating for $V$, then $\langle W, V, Y \rangle$ is a (shielded) collider (See Figure 3) if and only if every set that m-separates $X$ and $Y$ excludes $V$; it is a non-collider if and only if every set that m-separates $X$ and $Y$ contains $V$. The following proposition is proved in Spirtes and Richardson (1996)[1].

**Proposition 1.** *Two DMAGs over the same set of vertices are Markov equivalent if and only if*

*(e1) They have the same adjacencies;*

*(e2) They have the same unshielded colliders;*

*(e3) If a path u is a discriminating path for a vertex B in both graphs, then B is a collider on the path in one graph if and only if it is a collider on the path in the other.*

## 2 TRANSFORMATION BETWEEN EQUIVALENT DMAGS

We present the main result of the paper in this section, namely Markov equivalent DMAGs can be transformed to each other by a sequence of single mark changes that preserve Markov equivalence. We first describe in section 2.1 two corollaries from Zhang and Spirtes (2005) and Ali et al. (2005) which our arguments will rely upon. Section 2.2 establishes sufficient and necessary conditions for a single mark change to preserve Markov equivalence. The theorems are then presented in section 2.3.

### 2.1 LOYAL EQUIVALENT GRAPH

Given a MAG $\mathcal{G}$, a mark (or edge) in $\mathcal{G}$ is **invariant** if it is present in all MAGs Markov equivalent to $\mathcal{G}$. Invariant marks are particularly important for causal inference because data alone usually cannot distinguish between members of a Markov equivalence class. An algorithm for detecting all invariant arrowheads in a MAG is given by Ali et al. (2005), and one for further detecting all invariant tails is presented in Zhang and Spirtes (2005). The following is a special case of Corollary 18 in Zhang and Spirtes (2005).

**Proposition 2.** *Given any DMAG $\mathcal{G}$, there exists a DMAG $\mathcal{H}$ Markov equivalent to $\mathcal{G}$ such that all bi-directed edges in $\mathcal{H}$ are invariant, and every directed edge in $\mathcal{G}$ is also in $\mathcal{H}$.*

We will call $\mathcal{H}$ in Proposition 2 a **Loyal Equivalent Graph (LEG)** of $\mathcal{G}$. In general a DMAG could have multiple LEGs. A distinctive feature of the LEGs is that they have the fewest bi-directed edges among Markov equivalent DMAGs[2]. Drton and Richardson (2004) explored the statistical significance of this fact for fitting bi-directed graphs.

Another feature which will be particularly relevant to our argument is that between a DMAG and any of its LEGs, only one kind of difference is possible, namely, some bi-directed edges in the DMAG are oriented as directed edges in its LEG. For a simple illustration, compare the graphs in Figure 4, where H1 is a LEG of G1, and H2 is a LEG of G2. This feature is important because it will be the condition for Theorem 1 in 2.3.

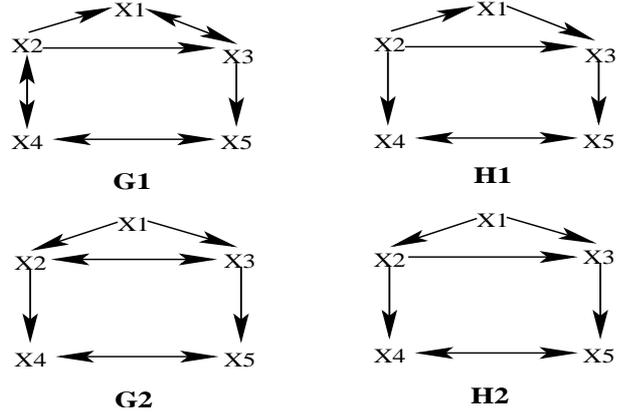

Figure 4: A LEG of G1 (H1) and a LEG of G2 (H2)

A directed edge in a DMAG is called **reversible** if there is another Markov equivalent DMAG in which the direction of the edge is reversed. To prove Theorem 2 in 2.3, we also need a fact that immediately follows from Corollary 4.1 in Ali et al. (2005).

**Proposition 3.** *Let $A \to B$ be any reversible edge in a DMAG $\mathcal{G}$. For any vertex C (distinct from A and B), there is an invariant bi-directed edge between C and A if and only if there is an invariant bi-directed edge between C and B.*

In particular, if $\mathcal{H}$ is a LEG of a DMAG, then $A \to B$ being reversible implies that A and B have the same spouses, as every bi-directed edge in $\mathcal{H}$ is invariant.

### 2.2 LEGITIMATE MARK CHANGE

Eventually we will show that any two Markov equivalent DMAGs can be connected by a sequence of equivalence-preserving mark changes. It is thus desirable to give some relatively simple graphical conditions under which a single mark change would preserve equivalence. Lemma 1 below gives necessary and sufficient conditions under which adding an arrowhead to a directed edge (i.e., changing the directed edge to a bi-directed one) preserves Markov equivalence. By symmetry, they are also the conditions for dropping

---

[1] The conditions are also valid for maximal ancestral graphs that contain undirected edges.

[2] For general MAGs, Corollary 18 in Zhang and Spirtes (2005) also asserts that the LEGs have the fewest undirected edges as well.

an arrowhead from a bi-directed edge while preserving Markov equivalence.

**Lemma 1.** *Let $\mathcal{G}$ be an arbitrary DMAG, and $A \rightarrow B$ an arbitrary directed edge in $\mathcal{G}$. Let $\mathcal{G}'$ be the graph identical to $\mathcal{G}$ except that the edge between $A$ and $B$ is $A \leftrightarrow B$. (In other words, $\mathcal{G}'$ is the result of simply changing $A \rightarrow B$ into $A \leftrightarrow B$ in $\mathcal{G}$.) $\mathcal{G}'$ is a DMAG and Markov equivalent to $\mathcal{G}$ if and only if*

*(t1) there is no directed path from $A$ to $B$ other than $A \rightarrow B$ in $\mathcal{G}$;*

*(t2) For any $C \rightarrow A$ in $\mathcal{G}$, $C \rightarrow B$ is also in $\mathcal{G}$; and for any $D \leftrightarrow A$ in $\mathcal{G}$, either $D \rightarrow B$ or $D \leftrightarrow B$ is in $\mathcal{G}$;*

*(t3) there is no discriminating path for $A$ on which $B$ is the endpoint adjacent to $A$ in $\mathcal{G}$.*

**Proof Sketch:** [3] We skip the demonstration of necessity because it is relatively easy and will not be used later. To prove sufficiency, suppose (t1)-(t3) are met, and we show that they guarantee $\mathcal{G}'$ is a DMAG and is Markov equivalent to $\mathcal{G}$. To see that $\mathcal{G}'$ is ancestral, note that it only differs from $\mathcal{G}$, an ancestral graph, regarding the edge between $A$ and $B$. So the only way for $\mathcal{G}'$ to violate the definition of ancestral graph is for $A$ to be an ancestor of $B$ in $\mathcal{G}'$, which contradicts (t1).

To show that $\mathcal{G}'$ is maximal, we need to show that there is no inducing path (Definition 4) between any two non-adjacent vertices. Suppose for contradiction that there is an inducing path $u$ in $\mathcal{G}'$ between two non-adjacent vertices, $D$ and $E$. Then $u$ includes $A \leftrightarrow B$ and $A$ is not an endpoint of $u$, for otherwise $u$ would also be an inducing path in $\mathcal{G}$, contradicting the fact that $\mathcal{G}$ is maximal. Also note that $D$ is not a parent of $B$, otherwise $D$ is an ancestor of $E$ by definition 4, which easily leads to a contradiction given that $\mathcal{G}'$ has been shown to be ancestral.

Suppose, without loss of generality, that $D$ is the endpoint closer to $A$ on $u$ than it is to $B$. Let $u(D, A) = \langle D = V_0, ..., V_n, A \rangle$ be the subpath of $u$ between $D$ and $A$. Then some $V_i$ is $B$'s spouse (in $\mathcal{G}$), for otherwise we can show by induction (starting from $V_n$) that every $V_i$, and in particular $V_0 = D$, is a parent of $B$, which is a contradiction.

Let $V_j$ be a spouse of $B$ on $u(D, A)$. Replacing $u(V_j, B)$ on $u$ with $V_j \leftrightarrow B$ yields an inducing path between $D$ and $E$ that does not include $A \leftrightarrow B$, and hence an inducing path between $D$ and $E$ in $\mathcal{G}$, a contradiction. So the initial supposition of non-maximality is false. $\mathcal{G}'$ is also maximal.

---
[3] The full version of the paper can be found at www.andrew.cmu.edu/user/jiji/transformation.pdf.

Lastly, we verify that $\mathcal{G}$ and $\mathcal{G}'$ satisfy the conditions for Markov equivalence in Proposition 1. Obviously they have the same adjacencies, and share the same colliders except possibly $A$. But $A$ will not be a collider in an unshielded triple, for condition (t2) requires that any vertex that is incident to an edge into $A$ is also adjacent to B. So the only worry is that a triple $\langle C, A, B \rangle$ might be discriminated by a path, but (t3) guarantees that there is no such path. $\square$

We say a mark change is *legitimate* when the conditions in Lemma 1 are satisfied. Recall that for DAGs the basic unit of equivalence-preserving transformation is (covered) edge reversal (Chickering 1995). In the current paper we treat an edge reversal as simply a special case of two consecutive mark changes. That is, a reversal of $A \rightarrow B$ is simply to first add an arrowhead at $A$ (to form $A \leftrightarrow B$), and then to drop the arrowhead at $B$ (to form $A \leftarrow B$). An edge reversal is said to be legitimate if both of the two consecutive mark changes are legitimate. Given Lemma 1, it is straightforward to check the validity of the following conditions for legitimate edge reversal. (We use $\mathbf{Pa}_\mathcal{G}/\mathbf{Sp}_\mathcal{G}$ to denote the set of parents/spouses of a vertex in $\mathcal{G}$.)

**Lemma 2.** *Let $\mathcal{G}$ be an arbitrary DMAG, and $A \rightarrow B$ an arbitrary directed edge in $\mathcal{G}$. The reversal of $A \rightarrow B$ is legitimate if and only if $\mathbf{Pa}_\mathcal{G}(B) = \mathbf{Pa}_\mathcal{G}(A) \cup \{A\}$ and $\mathbf{Sp}_\mathcal{G}(B) = \mathbf{Sp}_\mathcal{G}(A)$.*

When there is no bi-directed edge in $\mathcal{G}$, that is, when $\mathcal{G}$ is a DAG, the condition in Lemma 2 is reduced to the familiar definition for *covered edge*, i.e., $\mathbf{Pa}_\mathcal{G}(B) = \mathbf{Pa}_\mathcal{G}(A) \cup \{A\}$ (Chickering 1995). The condition given by Drton and Richardson (2004) for an edge in a bi-directed graph to be "orientable" in either direction ($\mathbf{Sp}_\mathcal{G}(B) = \mathbf{Sp}_\mathcal{G}(A)$) can be viewed as another special case of the above lemma.

### 2.3 THE MAIN RESULT

We first state two intermediate theorems crucial for the main result we are heading for. The first one says if the differences between two Markov equivalent DMAGs $\mathcal{G}$ and $\mathcal{G}'$ are all of the following sort: a directed edge is in $\mathcal{G}$ while the corresponding edge is bi-directed in $\mathcal{G}'$, then there is a sequence of legitimate mark changes that transforms one to the other. The second one says that if every bi-directed edge in $\mathcal{G}$ and every bi-directed edge in $\mathcal{G}'$ are invariant, then there is a sequence of legitimate mark changes (edge reversals) that transforms one to the other. The proofs follow the strategy of Chickering's proof for DAGs.

**Theorem 1.** *Let $\mathcal{G}$ and $\mathcal{G}'$ be two Markov equivalent DMAGs. If the differences between $\mathcal{G}$ and $\mathcal{G}'$ are all of the following sort: a directed edge is in $\mathcal{G}$ while the*

*corresponding edge is bi-directed in $\mathcal{G}'$, then there is a sequence of legitimate mark changes that transforms one to the other.*

**Proof Sketch:** We prove that there is a sequence of transformation from $\mathcal{G}$ to $\mathcal{G}'$, the reverse of which will be a transformation from $\mathcal{G}'$ to $\mathcal{G}$. Specifically we show that as long as $\mathcal{G}$ and $\mathcal{G}'$ are different, there is always a legitimate mark change that can eliminate a difference between them. The theorem then follows from a simple induction on the number of differences.

Let

$$\mathbf{Diff} = \{y|\ \text{there is a } x \text{ such that } x \rightarrow y \text{ is in } \mathcal{G} \text{ and } x \leftrightarrow y \text{ is in } \mathcal{G}'\}$$

By the antecedent condition, **Diff** exhausts all the differences there are between $\mathcal{G}$ and $\mathcal{G}'$. So the two graphs are identical if and only if $\mathbf{Diff} = \emptyset$. We claim that if **Diff** is not empty, there is a legitimate mark change that eliminates a difference. Choose $B \in \mathbf{Diff}$ such that no proper ancestor of $B$ in $\mathcal{G}$ is in **Diff**. Let

$$\mathbf{Diff}_B = \{x|x \rightarrow B \text{ is in } \mathcal{G} \text{ and } x \leftrightarrow B \text{ is in } \mathcal{G}'\}$$

Since $B \in \mathbf{Diff}$, $\mathbf{Diff}_B$ is not empty. Choose $A \in \mathbf{Diff}_B$ such that no proper descendant of $A$ in $\mathcal{G}$ is in $\mathbf{Diff}_B$. The claim is that changing $A \rightarrow B$ to $A \leftrightarrow B$ in $\mathcal{G}$ is a legitimate mark change — that is, it satisfies the conditions stated in Lemma 1.

The verifications of conditions (t1) and (t2) in Lemma 1 take advantage of the specific way by which we choose $A$ and $B$. For example, if condition (t1) were violated, i.e., if there were a directed path $d$ from $A$ to $B$ other than $A \rightarrow B$, then in order for $\mathcal{G}'$ to be ancestral, $d$ would not be directed in $\mathcal{G}'$, which implies that some edge on $d$ would be bi-directed in $\mathcal{G}'$. It is then easy to derive a contradiction to our choice of $A$ or $B$ in the first place. The verification of (t2) is similarly easy (which uses the fact that $\mathcal{G}$ and $\mathcal{G}'$ have the same unshielded colliders).

To show that (t3) also holds, suppose for contradiction that there is a discriminating path $u = \langle D, \cdots, C, A, B \rangle$ for $A$ in $\mathcal{G}$. By Definition 6, $C$ is a parent of $B$. It follows that the edge between $A$ and $C$ is not $A \rightarrow C$, for otherwise $A \rightarrow C \rightarrow B$ would be a directed path from $A$ to $B$, which has been shown to be absent. Hence the edge between $C$ and $A$ is bi-directed, $C \leftrightarrow A$ (because $C$, Definition 6, is a collider on $u$). Then the antecedent of the theorem implies that $C \leftrightarrow A$ is also in $\mathcal{G}'$. Moreover, the antecedent implies that every arrowhead in $\mathcal{G}$ is also in $\mathcal{G}'$, which entails that in $\mathcal{G}'$ every vertex between $D$ and $A$ is still a collider on $u$. It is then easy to prove by induction that every vertex between $D$ and $A$ on $u$ is also a parent of $B$ in $\mathcal{G}'$ (using the fact that $\mathcal{G}'$ is Markov equivalent to $\mathcal{G}$), and hence $u$ is also discriminating for $A$ in $\mathcal{G}'$. But $A$ is a collider on $u$ in $\mathcal{G}'$ but not in $\mathcal{G}$, which contradicts (e3) in Proposition 1. $\square$

Obviously a DMAG and any of its LEGs satisfy the antecedent of Theorem 1, so they can be transformed to each other by a sequence of legitimate mark changes. Steps 0-2, in Figure 5, for example, portraits a stepwise transformation from G1 to H1.

**Theorem 2.** *Let $\mathcal{G}$ and $\mathcal{G}'$ be two Markov equivalent MAGs. If every bi-directed edge in $\mathcal{G}$ is invariant and every bi-directed edge in $\mathcal{G}'$ is also invariant, then there is a sequence of legitimate mark changes that transforms one to the other.*

**Proof Sketch:** Without loss of generality, we prove that there is a transformation from $\mathcal{G}$ to $\mathcal{G}'$. Let

$$\mathbf{Diff} = \{y|\ \text{there is a } x \text{ such that } x \rightarrow y \text{ is in } \mathcal{G} \text{ and } x \leftarrow y \text{ is in } \mathcal{G}'\}$$

By the antecedent, **Diff** exhausts all the differences there are between $\mathcal{G}$ and $\mathcal{G}'$. If **Diff** is not empty, we can choose an edge $A \rightarrow B$ by exactly the same procedure as that in the proof of Theorem 1. The claim is that reversing $A \rightarrow B$ is a legitimate edge reversal (that is, a couple of legitimate mark changes), i.e., satisfies the conditions in Lemma 2.

The verification is fairly easy. Note that $A \rightarrow B$, by our choice, is a reversible edge in $\mathcal{G}$ (for $A \leftarrow B$ is in $\mathcal{G}'$, which is Markov equivalent to $\mathcal{G}$). It follows directly from Proposition 3 (and the assumption about bi-directed edges in $\mathcal{G}$) that $\mathbf{Sp}_{\mathcal{G}}(B) = \mathbf{Sp}_{\mathcal{G}}(A)$. The argument for $\mathbf{Pa}_{\mathcal{G}}(B) = \mathbf{Pa}_{\mathcal{G}}(A) \cup \{A\}$ is virtually the same as Chickering's proof for DAGs (Lemma 2, in particular, in Chickering 1995).

Note that after an edge reversal, no new bi-directed edge is introduced, so it is still true of the new graph that every bi-directed edge is invariant. Hence we can always identify a legitimate edge reversal to eliminate a difference in direction as long as the current graph and $\mathcal{G}'$ are still different. An induction on the number of differences between $\mathcal{G}$ and $\mathcal{G}'$ would complete the argument. $\square$

Since a LEG (of any MAG) only contains invariant bi-directed edges, two LEGs can always be transformed to each other via a sequence of legitimate mark changes according to the above theorem. For example, steps 2-4 in Figure 5 constitute a transformation from H1 (a

LEG of G1) to H2 (a LEG of G2). Note that Chickering's result for DAGs is a special case of Theorem 2, where bi-directed edges are absent.

We are ready to prove the main result of this paper.

**Theorem 3.** *Two DMAGs $\mathcal{G}$ and $\mathcal{G}'$ are Markov equivalent if and only if there exists a sequence of single mark changes in $\mathcal{G}$ such that*

1. *after each mark change, the resulting graph is also a DMAG and is Markov equivalent to $\mathcal{G}$;*

2. *after all the mark changes, the resulting graph is $\mathcal{G}'$.*

**Proof**: The "if" part is trivial – since every mark change preserves the equivalence, the end is of course Markov equivalent to the beginning. Now suppose $\mathcal{G}$ and $\mathcal{G}'$ are equivalent. We show that there exists such a sequence of transformation. By Proposition 2, there is a LEG $\mathcal{H}$ for $\mathcal{G}$ and a LEG $\mathcal{H}'$ for $\mathcal{G}'$. By Theorem 1, there is a sequence of legitimate mark changes $s_1$ that transforms $\mathcal{G}$ to $\mathcal{H}$, and there is a sequence of legitimate mark changes $s_3$ that transforms $\mathcal{H}'$ to $\mathcal{G}'$. By Theorem 2, there is a sequence of legitimate mark changes $s_2$ that transforms $\mathcal{H}$ to $\mathcal{H}'$. Concatenating $s_1$, $s_2$ and $s_3$ yields a sequence of legitimate mark changes that transforms $\mathcal{G}$ to $\mathcal{G}'$. □

As a simple illustration, Figure 5 gives the steps in transforming G1 to G2 according to Theorem 3. That is, G1 is first transformed to one of its LEGs, H1; H1 is then transformed to H2, a LEG of G2. Lastly, H2 is transformed to G2.

Theorems 1 and 2, as they are currently stated, are special cases of Theorem 3, but their proofs actually achieve a little more than they claim. The transformations constructed in the proofs of Theorems 1 and 2 are efficient in the sense that every mark change in the transformation eliminates a difference between the current DMAG and the target. So the transformations consist of as many mark changes as the number of differences at the beginning. By contrast, the transformation constructed in Theorem 3 may take some "detours", in that some mark changes in the way actually increase rather than decrease the difference between $\mathcal{G}$ and $\mathcal{G}'$. (This is not the case in Figure 5, but if, for example, we chose different LEGs for G1 or G2, there would be detours.) We believe that no such detour is really necessary, that is, there is always a transformation from $\mathcal{G}$ to $\mathcal{G}'$ consisting of as many mark changes as the number of differences between them. But we are yet unable to prove this conjecture.

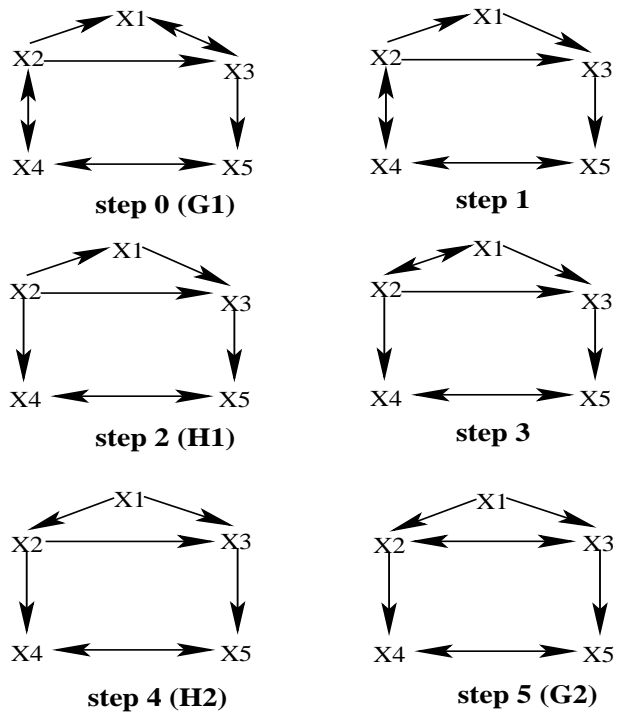

Figure 5: A transformation from G1 to G2

## 3 Conclusion

In this paper we established a transformational property for Markov equivalent directed MAGs, which is a generalization of the transformational characterization of Markov equivalent DAGs given by Chickering (1995). It implies that no matter how different two Markov equivalent graphs are, there is a sequence of Markov equivalent graphs in between such that the adjacent graphs differ in only one edge. It could thus simplify derivations of invariance properties across a Markov equivalence class — in order to show two arbitrary Markov equivalent DMAGs share something in common, we only need to consider two Markov equivalent DMAGs with the minimal difference. Indeed, Chickering (1995) used his characterization to derive that Markov equivalent DAGs have the same number of parameters under the standard CPT parameterization (and hence would receive the same score under the typical penalized-likelihood type metrics). The discrete parameterization of DMAGs is currently under development[4]. We think our result will prove useful to show similar facts once the discrete parameterization is available.

---

[4]Drton and Richardson (2005) provide a parameterization for bi-directed graphs with binary variables, for which the problem of parameter equivalence does not arise because no two different bi-directed graphs are Markov equivalent.

The property, however, does not hold exactly for general MAGs, which may also contain undirected edges[5]. A simple counterexample is given in Figure 6. When we include undirected edges, the requirement of ancestral graphs is that the endpoints of undirected edges are of zero in-degree — that is, if a vertex is an endpoint of an undirected edge, then no edge is into that vertex (see Richardson and Spirtes (2002) for details). So, although the two graphs in Figure 6 are Markov equivalent MAGs, M1 cannot be transformed to M2 by a sequence of single legitimate mark changes, as adding any single arrowhead to M1 would make it non-ancestral. Therefore, for general MAGs, the transformation may have to include a stage of changing the undirected subgraph to a directed one in a wholesale manner.

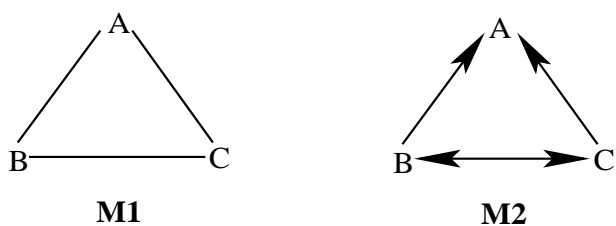

Figure 6: A simple counterexample with general MAGs: M1 can't be transformed into M2 by a sequence of legitimate single mark changes.

The transformational characterization for Markov equivalent DAGs was generalized, as a conjecture, to a transformational characterization for DAG I-maps by Meek (1996), which was later shown to be true by Chickering (2002). A graph is an I-map of another if the set of conditional independence relations entailed by the former is a subset of the conditional independence relations entailed by the latter. This generalized transformational property is used to prove the asymptotic correctness of the GES algorithm, an efficient search algorithm over the Markov equivalence classes of DAGs. The extension of both the property and the GES algorithm to MAGs is now under our investigation.

## Acknowledgement

We thank Thomas Richardson and Clark Glymour for reading an earlier draft of the paper, and anonymous referees for valuable comments.

---

[5]Undirected edges are motivated by the need to represent the presence of selection variables, features that influence which units are sampled (that are conditioned upon in sampling).

## References

Ali, R.A., T. Richardson, and P. Spirtes (2004) Markov Equivalence for Ancestral Graphs. Department of Statistics, University of Washington, Tech Report 466.

Ali, R.A., T. Richardson, P. Spirtes, and J. Zhang (2005) Towards Characterizing Markov Equivalence Classes of Directed Acyclic Graphs with Latent Variables. UAI 2005.

Andersson, S., D. Madigan, and M. Pearlman (1997) A Characterization of Markov Equivalence Classes of Acyclic Digraphs, in *Ann. Statist.* 25(2): 505-541.

Chickering, D.M. (1995) A transformational characterization of equivalent Bayesian network structures, in *Proceedings of Eleventh Conference on Uncertainty in Artificial Intelligence*, 87-98. Morgan Kaufmann.

Chickering, D.M. (2002) Optimal Structure Identification with Greedy Search, in *Journal of Machine Learning Research*, 3:507-554.

Drton, M., and T. Richardson (2004) Graphical Answers to Questions About Likelihood Inference in Gaussian Covariance Models. Department of Statistics, University of Washington, Tech Report 467.

Drton, M., and T. Richardson, T. (2005) Binary Models for Marginal Independence. Department of Statistics, University of Washington, Tech Report 474.

Meek, C. (1996) Graphical Models: Selecting Causal and Statistical Models. Carnegie Mellon University, Philosophy Department, PhD Thesis.

Richardson, T., and P. Spirtes (2002) Ancestral Graph Markov Models, in *Ann. Statist.* 30(4): 962-1030.

Spirtes, P., and T. Richardson (1996) A Polynomial Time Algorithm For Determining DAG Equivalence in the Presence of Latent Variables and Selection Bias, in *Proceedings of the 6th International Workshop on Artificial Intelligence and Statistics*.

Spirtes, P., T. Richardson, and C. Meek (1997) Heuristic Greedy Search Algorithms for Latent Variable Models, in *Proceedings of the 6th International Workshop on Artificial Intelligence and Statistics*.

Verma, T., and J. Pearl (1990) Equivalence and Synthesis of Causal Models, in *Proceedings of 6th Conference on Uncertainty in Artificial Intelligence*, 220-227.

Zhang, J., and P. Spirtes (2005) A Characterization of Markov Equivalence Classes for Ancestral Graphical Models. Department of Philosophy, Carnegie Mellon University, Tech Report 168, available at www.andrew.cmu.edu/user/jiji/completeness.pdf.